\documentclass{article}
\usepackage{spconf,amsmath,graphicx}
\usepackage{sidecap}
 \usepackage{tikz}
 \usepackage{makecell}
\usepackage{multirow}
\usepackage{hyperref}
\hypersetup{colorlinks=true}
\usepackage{cite}
\usepackage{enumitem}

\usetikzlibrary{decorations.pathreplacing}
 
\def\cat{{\tt Concat}}
\def\film{{\tt FiLM}}
\setlist[itemize]{leftmargin=*}

\title{Locale Encoding for scalable multilingual keyword spotting models}
\name{Pai Zhu, Hyun Jin Park, Alex Park, Angelo Scorza Scarpati, Ignacio Lopez Moreno }
\address{Google LLC,
Mountain View, CA, U.S.A\\
\texttt{\{paizhu,hjpark,axpk,angelos,elnota\}@google.com}}
\begin{document}
\ninept
\maketitle
\begin{abstract}
A Multilingual Keyword Spotting (KWS) system detects spoken keywords over multiple locales. Conventional monolingual KWS approaches do not scale well to multilingual scenarios because of high development/maintenance costs and lack of resource sharing. To overcome this limit, we propose two locale-conditioned universal models with locale feature concatenation and feature-wise linear modulation ($\film$). We compare these models with two baseline methods: locale-specific monolingual KWS, and a single universal model trained over all data. Experiments over 10 localized language datasets show that locale-conditioned models substantially improve accuracy over baseline methods across all locales in different noise conditions. $\film$ performed the best, improving on average FRR by 61\% (relative) compared to monolingual KWS models of similar sizes.

\end{abstract}
\section{Introduction}
\label{sec:intro}
Production-grade keyword-spotting (KWS) systems are trained to recognize keywords from a continuous stream of speech.  They operate in a resource-constrained, noisy environment.

Previously, research on this problem has focused on issues like noise robustness, reducing dependency on data volume and label quality, minimizing computing cost, and improving detection accuracy~\cite{Cnn15,Alvarez2019,MaxPool20,KwsStce21, AlexaMulti16,Progressive21, HeySiri17,UnifiedSpeculation22,EgoNoise22, Yang2022DeepRS, Hard2022ProductionFK, Gharbieh2022DyConvMixerDC}.  Most of the cited research addresses keyword spotting in a single specific language (locale). But for production-grade systems, it is highly important to scale up systems to support numerous international languages.  Similar to efforts in multilingual speech recognition~\cite{li2022massively} and multilingual speaker recognition~\cite{chojnacka2021speakerstew}, a universal multilingual KWS model will not only drastically reduce the cost of training, but also largely simplifies the model deployment process and maintenance cost.

In this paper, we discuss and explore a scalable approach to creating KWS models that cover numerous international languages while minimizing the cost of development at reasonable quality.

For the multilingual keyword spotting problem, we want to develop localized model(s) which can detect a desired keyword in various languages.  A naïve approach is to just develop a monolingual KWS model, and repeat the same process for other languages, simply switching out the training data and localized keyword.
This process will yield a set of $N$ locale specific models given $N$ locales.  It can serve as a simple baseline, with the drawback of having high maintenance costs and limited use of shared linguistic information across all of the training data.  Locale-specific data pre-processing and model training costs scale linearly with the number of locales, which can be prohibitive for tens or hundreds of locales.  Moreover, many common properties of acoustic data are likely helpful across all locales, and are not exploited by the repeated monolingual approach.



To overcome these limitations, we consider three new approaches for sharing information between locales, while minimizing development costs relative to the baseline. First, we consider a fully universal model, which is a single model trained with union of data from all locales.  Second, we propose two locale-conditioned universal models based on different conditioning methods: $\cat$ and $\film$. With the $\cat$ method, a locale encoding is  concatenated to the output of intermediate layer connecting encoder and decoder. With the $\film$ (Feature-wise  Linear  Modulation)~\cite{perez2018film} method, a locale encoding is used to modulate the same intermediate layer output. A locale conditioned universal model is a single model trained with all locales data together, but requiring the locale identity as an auxiliary input.  We compare these locale conditioned multilingual KWS methods with monolingual individual KWS method and fully shared universal model.
The rest of the paper is organized as follows: we discuss related works in Section~\ref{sec:related_work}, and describe the proposed approach in Section~\ref{sec:proposed_method}. We discuss the experimental setup and the result in Sections~\ref{sec:Experimental_setup} and~\ref{sec:Results}, and conclude in Section \ref{sec:conclusion}.

\section{Related Work and Background}
\label{sec:related_work}

\begin{figure*}[!htbp]
	\centering
	\includegraphics[width=0.95\textwidth]{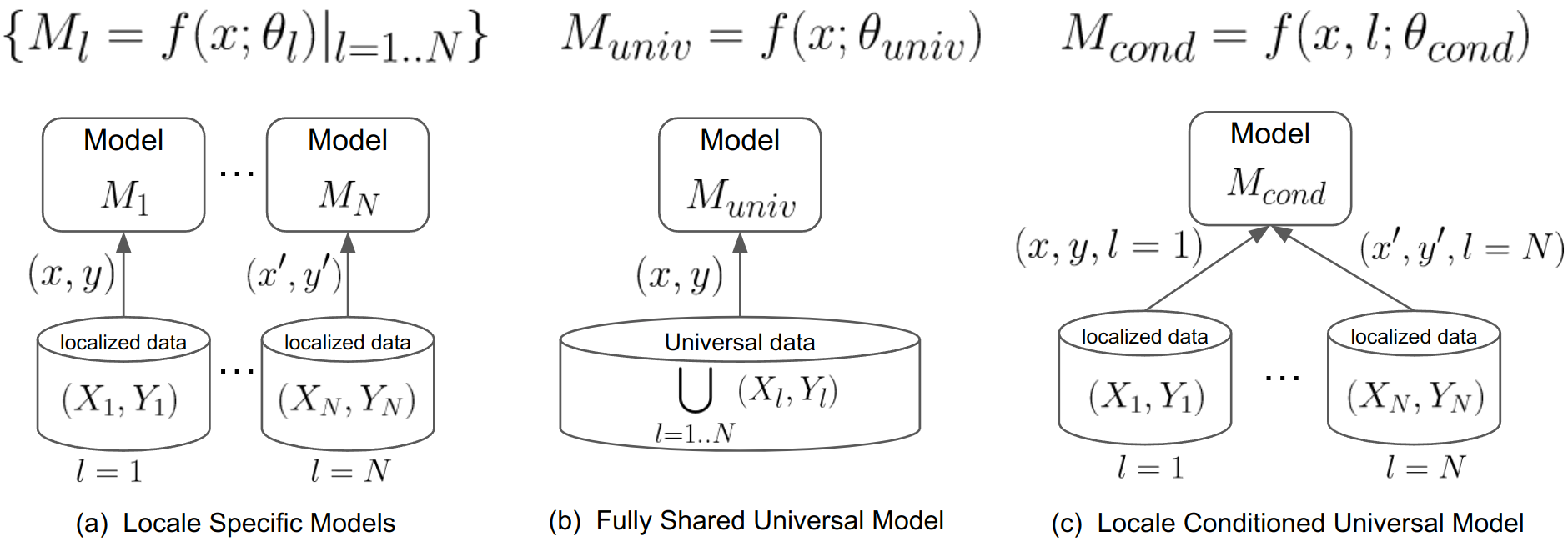}
	\caption{Different approaches of multilingual keyword spotting compared in this paper}
	\label{fig:three_approaches}
\end{figure*}

\subsection{Related Work} 

In~\cite{Multilingual2010}, multilingual KWS was explored by merging acoustically similar phonemes from two languages, and building a shared phoneme encoder based on HMM-NN. In~\cite{Menon2018FastAA,Menon2019FeatureEF}, a bottleneck feature encoder was trained to detect the union of phonemes from multiple languages to address multilingual KWS. More recently, \cite{Awasthi2021TeachingKS} showed that multilingual KWS models can benefit from learning shared embedding features. \cite{FewShot21} also trained an embedding model using multilingual data which is shown to generalize to unseen languages in few-shot setup. A common theme in these related works was the idea of learning shared representations that can generalize across locales, but not explicitly providing the locale information as an input. In~\cite{Toshniwal2018MultilingualSR,Kannan2019LargeScaleMS}, the authors showed that conditioning an ASR model with a one-hot locale encoding information is highly effective for multilingual generalization of the ASR model. With this work serving as inspiration, we proposes new locale-conditioned KWS approaches, which use shared parameters across different locales and also allow conditioning by an auxiliary locale encoding input.

\subsection{Baseline Model Architecture}
\label{sec:baselineArch}
For all models in this paper, we use an encoder-decoder architecture consisting of an encoder network (4 convolutional layers) followed by a decoder network (3 convolutional layers)~\cite{Alvarez2019,MaxPool20,KwsStce21}.  Note the convolution layers here are simplified versions described more fully in~\cite{Svdf15,Alvarez2019}. We refer to the connection between encoder and decoder networks as the bottleneck layer, which projects the final encoder convolution layer output to a dimension matching the input to decoder network.  We use $P$ to describe the encoder logits which are the outputs of this bottleneck layer.

The baseline encoder-decoder model is trained in a supervised manner, with training examples $(x,y)$ where $x$ is a sequence of input spectral feature vectors, and $y$ is a sequence of target labels for encoder and decoder logits.  We use cross-entropy loss following \cite{Alvarez2019} for the baseline and proposed models.

\section{Proposed Method}

\begin{figure*}[ht]
	\centering
	\includegraphics[width=0.15\textwidth]{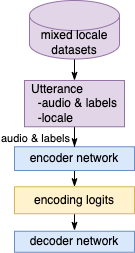}
	\hspace{1cm}
	\includegraphics[width=0.26\textwidth]{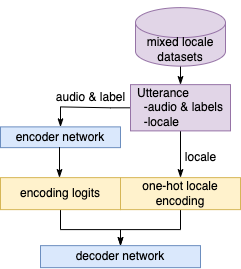}
	\hspace{1cm}
	\includegraphics[width=0.4\textwidth]{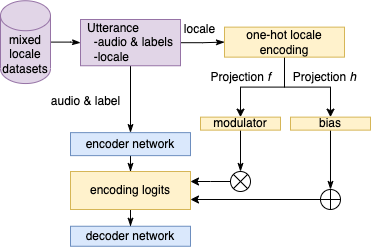}
	\caption{Fully shared universal model (left), locale concatenation model architecture (middle) and locale FiLM model architecture (right)}
	\label{fig:three_model_archs}
\end{figure*}


\label{sec:proposed_method}
Fig.\ref{fig:three_approaches} summarizes 3 different scaling approaches to multilingual keyword spotting problem.
Fig.\ref{fig:three_approaches} (a) shows a simple repetition approach where we repeat training of $N$ individual models using localized data for each locale.
Fig.\ref{fig:three_approaches} (b) shows a fully shared model approach where we train a single model with training data from all locales.
Fig.\ref{fig:three_approaches} (c) shows a locale conditioned model approach where we train a single model with training data from all locales and corresponding locale information. Throughout this paper, we denote $(X, Y)$ as a composite sequence defined as  $( (x_i,y_i)|i=1..k )$,  given original feature sequence $X=( x_i  | i=1..k )$ and label sequence $Y=( y_i | i=1..k)$ of length $k$. $X_l$ and $Y_l$ denote feature and label sequences from locale $l$ respectively.

\subsection{Locale Specific Models}
A locale specific model for locale $l$ can be defined as $M_l = f(x;\;\theta_l)$ where $x$ is the input features, and $\theta_l$ is the set of trainable parameters for each $M_l$ (See Fig.\ref{fig:three_approaches}-a). Such models can be trained by minimizing the expected losses per each model,
\begin{equation}
\label{eq:locale_specific_model}
\begin{aligned}
\theta_l \; = & \; \mathtt{argmin} \: E_{(x,y)}[\mathit{Loss}(f(x;\;\theta_l),y)] \\
           & \mathtt{where} \: (x,y) \in (X_l, Y_l) \:,\: l \in {\left \{1..N \right \}}
\end{aligned}
\end{equation}
Here, we use $E_{\ast}[\ldots]$ to denote expectation over $\ast$.

\subsection{Fully Shared Universal Model}
A fully shared universal model can be defined as $M_{\tt univ}= f(x;\;\theta_{\tt univ})$ where $\theta_{\tt univ}$ is a single set of parameters shared across all locales (see Fig.\ref{fig:three_approaches}-b).
In this case, there will be only a single model trained on the pooled data,

\begin{equation}
\label{eq:fully_universal_model}
\begin{aligned}
\theta_{\tt univ} \; = & \; \mathtt{argmin} \: E_{(x,y)}[\mathit{Loss}(f(x;\;\theta_{\tt univ}),y)] \\
           & \mathtt{where} \: (x,y) \in \bigcup_{l=1..N} (X_l , Y_l)
\end{aligned}
\end{equation}

\subsection{Locale Conditioned Universal Model}
\label{ssec:localeSection}
A locale conditioned model can be described similarly to the previously defined universal model, $M_{\tt cond} = f(x,l;\theta_{\tt cond})$, except for the additional locale encoding input, $l$.  As with $\theta_{\tt univ}$, the $\theta_{\tt cond}$ are shared across all locales (see  Fig.\ref{fig:three_approaches}-c).
\begin{equation}
\label{eq:locale_conditioned_model}
\begin{aligned}
\theta_{\tt cond} = & \mathtt{argmin} \:\: E_{(x,y,l)}[\textit{Loss}(f(x,l;\theta_{\tt cond}),y)] \\
           & \mathtt{where} \: (x,y) \in \: (X_l, Y_l),\:  l \in \{1..N\}
\end{aligned}
\end{equation}
We experiment with two methods for locale conditioning: concatenation ($\cat$), and modulation ($\film$).
\subsubsection{Concatenation}
\label{sssec:concat}

 As shown in Fig.\ref{fig:three_model_archs} (middle), each training example (i.e. utterance) comes with a locale index, $l \in \{1..N\}$, denoting the utterance locale origin. We represent locales as one-hot vectors, $L$, with length $N$. The locale index position will have value 1 and elsewhere are 0.  With the $\cat$ approach, we simply concatenate $L$ with the encoder logits, $P$, and use the resulting combined tensor as the input to the decoder network.  The extra size introduced to the model is $N \times D_1$ where $D_1$ is the first decoder layer input dimension.

\subsubsection{Modulation}
Another approach to conditioning is to use the locale information to modulate the existing encoder logits.  Feature-wise Linear Modulation ($\film$)~\cite{perez2018film} learns to adaptively influence the neural network output by applying an affine transform to the network's intermediate features based on an external input. In this case, the external input is the one-hot encoded locale $L$ as defined in Section~\ref{sssec:concat}, and the intermediate features are the above mentioned encoder logits $P$.

Formally, $\film$ learns element-wise modulation (scale) and bias (shift) functions which can be implemented as simple learnable projection layers, $f$ and $h$, respectively. As shown in Fig.~\ref{fig:three_model_archs} (right), $f$ projects $L$ to create a modulation factor, $\gamma$, and $h$ projects $L$ to create a bias factor, $\beta$.  Both $\gamma$ and $\beta$ have the same dimension as $P$, the encoder logits. The conditioning is then applied on $P$ to produce the modulated input to the decoder network,

\begin{equation}
P^{\tt mod} \; = \; \film(P|\gamma, \beta) \; = \; \gamma \cdot P + \beta
\end{equation}

where $\cdot$ denotes elementwise multiplication.  The number of extra parameters introduced with $f$ and $h$ is $2 \times M \times N$ where $M$ is the size of encoder logits and $N$ is the number of locales.

\section{Experimental setup}
\label{sec:Experimental_setup}

\subsection{Train and Eval Datasets}
\label{sec:train_and_eval_dataset}

Our dataset consists of 1.2 billion anonymized utterances from real-world queries collected  in  accordance  with  Google  Privacy and AI  principles~\cite{privacyprinciples, aiprinciples}. Utterances containing localized versions of keyword phrase ``Ok Google'' or ``Hey Google'' are referred to as positive data. Negative utterances are collected from queries triggered by tactile (push-button) and hence do not contain the keyword phrases.

Data is divided into training and evaluation sets as in~\cite{KwsStce21}. The training set consists of 838M positive and 435M negative utterances. The training set has been supplemented with different transformations and noises producing 25 new replicas for each utterance~\cite{Alvarez2019, Kim2017}. The evaluation set contains 6M positive utterances divided into 5M regular positive utterances and 1M challenging positive utterances based on SNR. The evaluation set also contains 5k hours of negative audio used to determine the operating point threshold needed for a given FA/hour level.





\subsection{Metrics}
\label{ssec:metrics_def}
We evaluate False Reject Rate (FRR), which is the number of positive utterances rejected by the model divided by the number of positive utterances. Similarly we measure False Acceptance Per Hour (FAh) which is the number of false detections divided by the number of negative audio hours. We measure FRR in positive sets with various noise conditions and FAh in negative sets at various threshold levels as mentioned in Section~\ref{sec:train_and_eval_dataset}. We compare model performances by FRR values at a consistent FAh level (0.17/hour) and their FRR-FAh curve plots with the area of interest 0-0.5 FAh.

\begin{table*}[!htbp]
\centering
    \resizebox{18cm}{!}{
    \begin{tabular}{l l|r|r|r|r|r|r|r|r|r|r|r}
     \multirow{6}{*}{Eval-reg}
        & Locale-name & DA\_DK & DE\_DE & ES\_ES & FR\_FR & IT\_IT & KO\_KR & NL\_NL & PT\_BR & SV\_SE & TH\_TH & AVERAGE \\
    \Xhline{4\arrayrulewidth}
    & Locale Specific Models  & 22.61  & 2.49  & 5.67  & 8.47  & 6.09  & 11.57  & 5.81  & 6.92  & 19.35  & 3.10  & 9.21  \\ \cline{2-13}
        & Universal Model    &  6.15  & 4.66  & 10.89  & 10.66  & 6.34  & 10.05  & 4.69  & 7.58  & 14.89  & 6.64  & 8.26  \\ 
    \cline{2-13}
        & Locale $\cat$ Model & 1.92  & 2.97  & 11.16  & 14.07  & 3.95  & 5.29  & 1.18  & 2.53  & 4.79  & 3.13  & 5.10  \\
    \cline{2-13}
        & Locale $\film$ Model & 1.62  & 2.16  & 4.65  & 7.26  & 3.25  & 6.29  & 2.58  & 2.38  & 4.20  & 1.66  & \textbf{3.60 } \\
    \Xhline{4\arrayrulewidth}
    \multirow{4}{*}{Eval-chall}
    &  Locale Specific Models & 51.19  & 24.65  & 28.84  & 38.45  & 29.75  & 55.52  & 19.57  & 31.99  & 52.60  & 10.09  & 34.26  \\ \cline{2-13}
        & Universal Model & 44.05  & 45.17  & 48.73  & 50.12  & 31.38  & 46.33  & 20.37  & 35.92  & 48.64  & 34.20  & 40.49  \\ \cline{2-13}
        & Locale $\cat$ Model & 16.39  & 28.47  & 40.81  & 43.75  & 20.41  & 31.55  & 6.21  & 13.41  & 24.83  & 16.25  & 24.21  \\ \cline{2-13}
        & Locale $\film$ Model & 12.33  & 17.17  & 21.82  & 20.65  & 16.37  & 30.47  & 10.52  & 13.51  & 20.72  & 8.53  & \textbf{17.21 } \\
    \end{tabular}}
    
    \caption{FRRs (\%) of different models in 10 locale datasets with regular (Eval-reg) and challenging (Eval-chall) acoustic conditions, from the 330K params model $M_R$. The thresholds are chosen to have the same targeted FAh (0.17) in the negative audio set.}
    \label{tab:FRR_Table_Small_Model}
\end{table*}

\begin{table*}[!htbp]
\centering
    \resizebox{18cm}{!}{
    \begin{tabular}{l l|r|r|r|r|r|r|r|r|r|r|r}
     \multirow{6}{*}{Eval-reg}
        & Locale-name & DA\_DK & DE\_DE & ES\_ES & FR\_FR & IT\_IT & KO\_KR & NL\_NL & PT\_BR & SV\_SE & TH\_TH & AVERAGE \\
    \Xhline{4\arrayrulewidth}
        & Locale Specific Models & 9.74 & 2.58 & 4.63 & 7.09 & 4.72 & 7.97 & 8.59 & 6.66 & 7.80 & 6.70 & 6.65 \\ \cline{2-13}
        &  Universal Model & 3.59 & 3.29 & 7.47 & 8.13 & 4.27 & 6.23 & 2.19 & 4.31 & 8.44 & 4.87 & 5.28 \\ \cline{2-13}
        & Locale Concat Model & 1.64 & 2.03 & 5.41 & 6.31 & 2.78 & 7.32 & 0.85 & 2.25 & 4.51 & 1.76 & 3.49 \\ \cline{2-13}
        & Locale FiLM Model & 1.94 & 1.51 & 5.81 & 4.93 & 1.93 & 4.50 & 1.02 & 1.79 & 3.37 & 1.53 & \textbf{2.83} \\

    \Xhline{4\arrayrulewidth}
    \multirow{4}{*}{Eval-chall}
    & Locale Specific Models & 28.84 & 25.76 & 25.94 & 34.68 & 24.80 & 46.91 & 22.53 & 31.38 & 25.84 & 22.31 & 28.90 \\ \cline{2-13}
        &  Universal Model & 28.54 & 30.79 & 37.61 & 32.11 & 22.90 & 32.20 & 11.29 & 21.26 & 32.96 & 24.45 & 27.41 \\ \cline{2-13}
        & Locale Concat Model & 14.65 & 16.27 & 22.61 & 20.62 & 13.72 & 28.94 & 3.80 & 11.09 & 19.40 & 9.35 & 16.04 \\ \cline{2-13}
        & Locale FiLM Model & 14.05 & 14.05 & 23.68 & 17.96 & 10.92 & 23.09 & 3.83 & 9.48 & 18.15 & 7.24 & \textbf{14.24} \\
    \end{tabular}}
    \caption{FRRs (\%) of different models in 10 locale datasets with regular (Eval-reg) and challenging (Eval-chall) acoustic conditions, from the 1.4M params model $M_L$. The thresholds are chosen to have the same targeted FAh (0.17) in the negative audio set.}
    \label{tab:FRR_Table_Large_Models}
\end{table*}

\subsection{Training Details}

\begin{itemize}
\item  \textbf{Locales}: The locales in our experiments include : DA-DK(Danish), DE-DE (German), ES-ES (Spanish), FR-FR (French), IT-IT (Italian), KO-KR (Korean), NL-NL (Dutch), PT-BR (Brazilian Portuguese), SV-SE (Swedish), TH-TH (Thai).

\item  \textbf{Target keywords}: Localized versions of `OK Google' and `Hey Google'.

\item  \textbf{Train environment}: Tensorflow/Lingvo \cite{Shen2019LingvoAM} is used.

\item  \textbf{Input feature and labels}: Stacked 40d spectral energy is used as the input feature $x$. Target labels $y$ are derived from force-alignment as described in \cite{Alvarez2019}. We merge similarly sounding phonemes over different languages for encoder labels.

\item    \textbf{Train loss/steps}: Cross entropy is used as loss for encoder and decoder following \cite{Alvarez2019}. We train for 3M$\sim$8M steps until loss converges to a stable level.

\item  \textbf{Model size}: We experiment with various model sizes: Regular ($M_R$) with 330K parameters \cite{Alvarez2019}, Large ($M_L$) with 1.4M parameters, and XLarge ($M_{XL}$) with 2.4M parameters.
\item  \textbf{Dimension of encoder logits}: $|P|$=32
\end{itemize}

\begin{figure*}[!ht]
	\centering
	\includegraphics[width=1.0\textwidth,height=147pt]{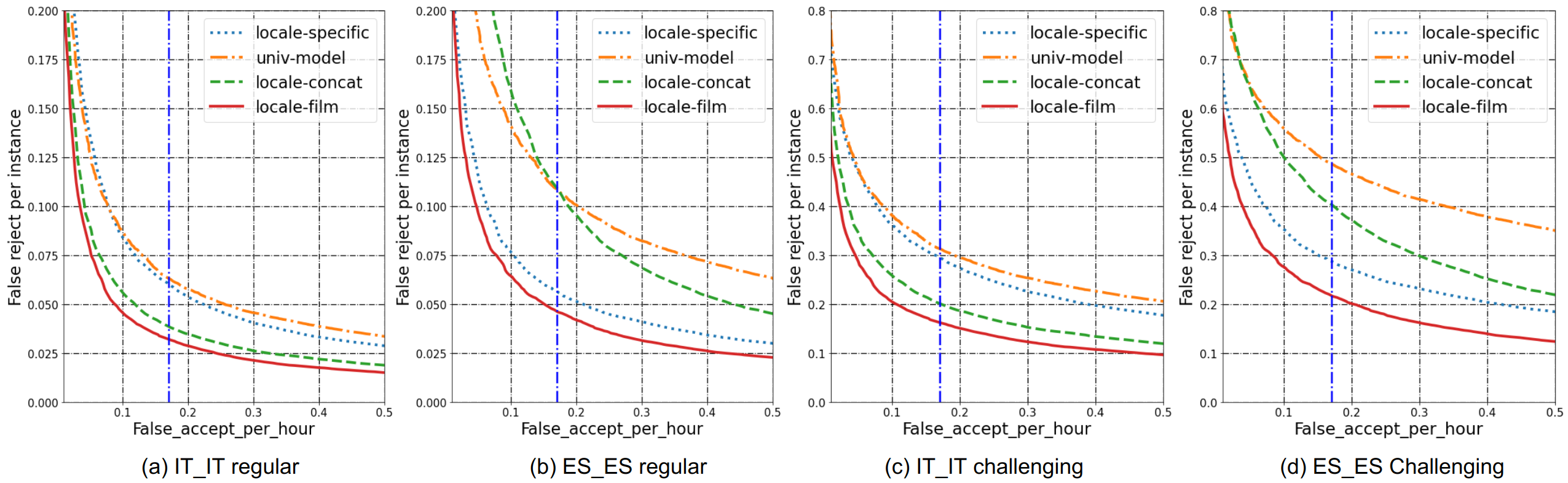}
	\caption{FRR-FAh plots for ES\_ES and IT\_IT under regular and challenging acoustic conditions for 330k params model $M_R$. The false reject per instance (FRR) and false accept per hour (FAh) are computed from positive and negative audio sets respectively in Section \ref{sec:train_and_eval_dataset}. Note the dotted blue vertical line represents the targeted 0.17 FAh and its intersections with curves show relevant FRR values in Table \ref{tab:FRR_Table_Small_Model}.}
	\label{fig:roc_plots}
\end{figure*}

\section{Results}
\label{sec:Results}

As discussed above, the baseline models include \textbf{locale specific models}, with each independently trained with their own locale data, and a \textbf{fully shared universal model} trained with mixed locale data. The experimental models integrated the locale encoding into the network in through $\cat$ and $\film$ approaches mentioned in Section~\ref{ssec:localeSection}. FRRs are reported based on the threshold determined at a constant FAh mentioned in Section~\ref{ssec:metrics_def}.

Table \ref{tab:FRR_Table_Small_Model} shows FRR results for the above models in both regular and challenging acoustic condition evaluation sets. We also plot FRR-FAh curves for selected locales in Fig \ref{fig:roc_plots}. Noticeably, Some locales have bad locale specific models due to the low volume or poor label quality of the data.  This is particularly true of smaller locales such as DA\_DK and SV\_SE, which have an order of magnitude of less training data than large locales such as ES\_ES and TH\_TH.  We observe improved metrics on these underrepresented locales with the universal model because the paucity of training data is compensated for by data from other locales, leading to less over-fitting.

%


For many other locales, the universal model has poorer results than the locale specific model, most likely because less relevant training data from other locales can harm specialization when there \textbf{is} sufficient data to get reasonable locale-specific performance. 

In both acoustic conditions, locale-conditioned models achieved significantly better results than locale-specific models and the universal model. On the one hand, the locale encoding models enhanced the training data volume from cross locale mixing. But unlike the universal model, the locale encoding networks is trained to selectively focus on data relevant to the locale whose input is provided.

We find that $\film$ consistently outperforms $\cat$, and conjecture that it is due to improved efficiency in learning locale similarities. In the concatenative approach, the locale encoding network is trained with the combined input consisting of encoder logits and locale encoding so that forward propagation is interfered undesirably. In $\film$, the locale encoding network takes only locale encoding as input and the learned weights can be used to compare locale similarities by calculating correlation matrix.

Large model $M_L$ gives enhanced learning capacity hence the locale specific model and universal model have boosted results in Table~\ref{tab:FRR_Table_Large_Models}. The locale encoding approaches further improved the results for different acoustic conditions. We also experimented with an even bigger size $M_{XL}$, but models in general have slight worse results than $M_L$, possibly due to over-fitting.

\section{Conclusion}
\label{sec:conclusion}
This paper introduced two new approaches for training multilingual KWS models - locale conditioned universal model with concatenation and $\film$ modulation approaches. Experiments show that both approaches significantly outperform locale-specific models and the fully shared universal model across 10 different language datasets in various acoustic conditions. Experiments with larger model sizes also show consistent improvements by the proposed approaches. The $\film$ approach achieves the best results given the efficient way learning cross-locale similarities. Result in Table \ref{tab:FRR_Table_Small_Model} shows that $\film$ based approach reduced average FRR by as much as 61\% relatively compared to locale specific models. The idea can be extended to other cross domain scenarios to utilize data efficiently when training.

\clearpage
\bibliographystyle{IEEEbib}
\bibliography{strings,refs}

\end{document}